\documentclass[preprint]{imsart}

\RequirePackage[OT1]{fontenc}
\RequirePackage[numbers]{natbib}

\usepackage{amsthm,amsmath}
\usepackage{amsbsy}

\input xy
\xyoption{all}

\usepackage{algorithm}
\usepackage{algpseudocode}

\usepackage{epsf}
\usepackage{epsfig}

\usepackage{enumitem}

\begin{document}

\begin{frontmatter}

% "Title of the Paper"
\title{Detecting Learning vs Memorization in Deep Neural Networks using Shared Structure Validation Sets}
%\thankstext{t1}{This work has been supported by grant \ldots}
\runtitle{Learning vs memorization in DNNs using s-validation sets}

% indicate corresponding author with \corref{}
% \author{\fnms{John} \snm{Smith}\thanksref{t2}\corref{}\ead[label=e1]{smith@foo.com}\ead[label=e2,url]{www.foo.com}}
% \thankstext{t2}{Thanks to somebody}
% \address{line 1\\ line 2\\ \printead{e1}\\ \printead{e2}}

%\author{\fnms{Elias} \snm{Chaibub Neto, Sage mPower Team}\ead[label=e1]{elias.chaibub.neto@sagebase.org, Sage Bionetworks}}
%\address{\printead{e1}}

\author{\fnms{} \snm{Elias Chaibub Neto$^{1}$}\ead[label=e1]{$^\ast$ elias.chaibub.neto@sagebase.org, $^1$ Sage Bionetworks}}
\address{\printead{e1}}

%$^\ast$\footnote{Corresponding author: elias.chaibub.neto@sagebase.org}

%\and
%\author{\fnms{???} \snm{???}\ead[label=e2]{???}}
%\address{\printead{e2}}

\runauthor{Chaibub Neto E.}
%\runauthor{}

\begin{abstract}
The roles played by learning and memorization represent an important topic in deep learning research. Recent work on this subject has shown that the optimization behavior of DNNs trained on shuffled labels is qualitatively different from DNNs trained with real labels. Here, we propose a novel permutation approach that can differentiate memorization from learning in deep neural networks (DNNs) trained as usual (i.e., using the real labels to guide the learning, rather than shuffled labels). The evaluation of weather the DNN has learned and/or memorized, happens in a separate step where we compare the predictive performance of a shallow classifier trained with the features learned by the DNN, against multiple instances of the same classifier, trained on the same input, but using shuffled labels as outputs. By evaluating these shallow classifiers in validation sets that share structure with the training set, we are able to tell apart learning from memorization. Application of our permutation approach to multi-layer perceptrons and convolutional neural networks trained on image data corroborated many findings from other groups. Most importantly, our illustrations also uncovered interesting dynamic patterns about how DNNs memorize over increasing numbers of training epochs, and support the surprising result that DNNs are still able to learn, rather than only memorize, when trained with pure Gaussian noise as input.
\end{abstract}

%\begin{keyword}[class=AMS]
%\kwd[Primary ]{}
%\kwd{}
%\kwd[; secondary ]{}
%\end{keyword}

%\begin{keyword}
%\kwd{}
%\kwd{}
%\end{keyword}

% history:
% \received{\smonth{1} \syear{0000}}

%\tableofcontents

\end{frontmatter}

\section{Introduction}

Recently, the roles of learning and memorization in deep neural networks (DNNs) have been investigated by several groups\cite{zhang2017,arpit2017}. Surprising empirical results by Zhang et al\cite{zhang2017} showed that DNNs are able to fit pure noise, establishing that the effective capacity of DNNs is sufficient for memorizing the entire data set, and raising the question whether memorization played a similar role in real data-sets. Further investigations by Arpit et al\cite{arpit2017}, showed, nonetheless, that there are qualitative differences on how DNNs learn when trained on real data compared to noisy versions of the data.

Intuitively, the term ``learning" captures the idea that the algorithm is able to discover patterns in the training inputs that are associated with the labels. ``Memorization", on the other hand, can be understood as the ability to discover patterns that are unrelated to the labels. Note that memorization can be directly assessed by evaluating the predictive performance of a trained algorithm on a data-set that shares some structure with the training set. For instance, Zhang et al\cite{zhang2017} were able to establish the memorization issue in DNNs by inspecting the training error, rather than the validation or test error, of models trained on shuffled labels.

While the label shuffling technique adopted by Zhang et al\cite{zhang2017}, is able to show beyond any doubts that DNNs can easily memorize, it does not provide insights about the interplay between memorization and learning in DNNs trained with the real labels. As shown by Arpit et al\cite{arpit2017}, DNNs behave differently when trained in noise vs real labels. In particular, their results suggest that DNNs trained on real labels tend to learn simple patterns first, before starting to memorize.

Here, we propose a novel permutation approach to assess the interplay between memorization and learning in DNNs trained with real labels (as opposed to shuffled labels). Its construction involves the generation of perturbed versions of the training data, denoted the ``shared structure validation sets" (or ``s-validation sets", for short), and the comparison of the predictive performance of a shallow classifier trained with the features learned by the DNN, against multiple instances of the same classifier, trained on the same input, but using shuffled labels as outputs. By evaluating these shallow classifiers in the s-validation sets, we are able to tell apart learning from memorization (as fully described in the next sections). Most importantly, because our assessments are based on the features learned by DNNs trained as usual, our approach can assess the memorization and learning processes that occur naturally during training (rather than assessing memorization under the more unusual setting of training with shuffled labels).

Our methodology can be used to compare how different combinations of network architectures, optimizers, and regularization techniques are able to favor or reduce memorization. We illustrate its application to multi-layer perceptrons (MLPs) and convolutional neural networks (CNNs), with and without dropout regularization\cite{dropout2014}, using the Fashion-MNIST data-set\cite{fashion2017}. Most importantly, we also investigate the role of learning and memorization with noise data by replacing the image's pixels by pure Gaussian noise, and evaluating the same architecture/regularization combinations used to fit the real data-sets.

While our main goal here is to introduce this novel tool (rather than provide an exhaustive empirical study comparing how DNNs trained with different combinations of architectures/regularizers/optimizers memorize and learn), our examples illustrate and corroborate the findings from Arpit et al\cite{arpit2017} that DNNs are able to memorize without hindering their ability to learn, and that regularization (in our case dropout) can effectively reduce memorization in DNNs. But, most interestingly, application of our method across a sequence of increasing training epochs show interesting dynamic patterns of how DNNs memorize over training time. Furthermore, our illustrations using noise inputs support the surprising result that that DNNs can still learn, rather than only memorize, when trained with noise inputs.

The implementation of our permutation approach involves a series of steps that we described in the detail in the next sections. But first, we provide more detailed (while still informal) working definitions of learning and memorization, and describe their connection to generalization.

\section{Learning vs memorization and their connection to generalization}

The goal of any supervised learning algorithm is to learn patterns in the input data that are associated with the concepts expressed by the labels. For instance, in image applications one might be interested in training an algorithm to distinguish images of dogs from images of cats, and the goal is to train a classifier that can learn patterns in the image pixels that are associated with the concept of a ``dog" versus the concept of a ``cat".

The ability to learn such patterns depends, of course, on the input data having some structure that is shared among images containing the same label concept. Throughout this text we refer to this shared structure as the \textit{label structure} of the data. Observe, nonetheless, that labeled images might contain additional structure associated with their background, that we denote the \textit{background structure} of the data\footnote{Observe that while these working ``definitions" of label and background structure are undeniably fuzzy, they are still helpful as instructive concepts. Note, as well, that one can still think about label and background structure concepts in other types of problems, including speech recognition, medical diagnostic applications, and etc.}.

%the same concepts are applicable not only to image classification, but also

Here, we adopt the operational ``definition" of \textit{learning} as the ability of a machine learning algorithm to discover patterns in the training inputs that are associated with the label concepts. In other words, learning boils down to discovering label structure in the training data. \textit{Memorization}, on the other hand, can be ``defined" as the ability to discover patterns in the training inputs that are unrelated to the label concepts. In other words, memorization can be equated to learning the background structure of the training data. (As will be explained in later sections, memorization can be assessed by evaluating classification performance in data-sets that shares background structure with the training data.)

Note that these definitions of learning and memorization are restricted to processes that happen during training, and only depend on the training data. Generalization, which corresponds to an ability to correctly predict the label of an unseen input sample (with a probability better than a random guess), represents a separate concept, and is all about recognizing label structures learned during training in independent validation and test sets.

In general, we do not expect that models that have only memorized the labels, will be able to generalize when evaluated in independent validation or test sets (because validation and test sets usually share label structure with the training set, but not a whole lot of background structure). Models that have learned the labels, on the other hand, will usually be able to generalize because they are able to recognize label structures in the validation and test sets that are similar to the label structure that they learned in the training set. Observe, however, that even the best trained algorithms won't be able to generalize if the validation and test sets do not share label structures with the training set.

\section{The shared structure validation set}

We use data perturbation to generate a validation set that shares structure with the training set (the s-validation set). More specifically, each sample of the training set is perturbed in order to generate a corresponding sample in the s-validation set (hence, training and s-validation sets have the exact same number of samples, and the same labels). For instance, in image applications, we can shuffle a fixed proportion of pixels in the training image, to generate the corresponding image in the s-validation set. (Alternatively, one could rotate the images in the training set to generate the images in the s-validation set.) Note that the s-validation set shares both background and label structure with the training set, while the validation and test sets share mostly label structure with the training set.

Clearly, by increasing the amount of perturbation, we can generate s-validation sets that share decreasing amounts of structure with the training set. (Note that by perturbing too much the training inputs, we risk whipping out the label structure of the training data, possibly making the predictive performance in the s-validation set worse than in the validation set.) In practice, we generate multiple s-validation sets using increasing amounts of perturbation, but starting with light perturbations. As will be made clear by our illustrative examples, the real use of our approach is to compare the relative amounts of memorization and learning across distinct DNNs, trained and evaluated in the same training and s-validation sets, rather than attempting to quantify the absolute amount of memorization happening in a DNN.

\section{Overall description of the procedure}

The implementation of our permutation approach involves five key steps. Here, we provide an overall description of the mechanics of the process:

\begin{enumerate}[label=\roman*.]

\item We generate the s-validation sets, as described in the previous section. In our experiments we use s-validation sets generated by shuffling 5\%, and 10\% of the pixels of the original training images. (We also generated a completely scrambled s-validation set, where we shuffled 100\% of the image pixels, in order to perform some sanity checks).

\item We fit a DDN to the training data. Note that the DNN is trained as usual, using the original (unshuffled) training labels, and is potentially able to learn and memorize.

\item We generate the training set of learned-features by propagating the raw input samples from the training set through the trained DNN, and extracting the feature values generated by the last hidden-layer of the DNN (before the output layer). Similarly, we generate the s-validation sets of learned-features by propagating the raw input samples from the s-validation sets through the trained DNN, and extracting the features from the last hidden-layer of the DNN. (We also generate validation sets of learned-features in a similar way. Even though this data is not used by our permutation approach, we still evaluate validation errors as sanity checks in our examples.)

\item Once, we have the training and s-validation sets of learned-features, we: fit a shallow machine learning classifier (e.g., random forest) to the training features; generate predictions using the s-validation features; and evaluate these predictions against the training labels (which correspond to the s-validation labels) using a classification performance metric (e.g., classification accuracy). We denote this performance metric value, the \textit{observed s-validation performance score}.

\item Finally, we generate a permutation null distribution to assess whether the DNN has learned and/or memorized using Algorithm \ref{alg:permnull}, described in the next section.

\end{enumerate}

\section{A permutation null distribution for detecting learning vs memorization in DNNs}

Once we have generated the training and s-validation sets of learned-features (according to step iii described above), and have computed the observed performance scores associated with each s-validation set (according to step iv above), we can generate permutation null distributions to tell apart learning from memorization in DNNs, as described in Algorithm \ref{alg:permnull}.

\begin{algorithm}[b]
   \caption{Learning vs memorization null distribution}
   \label{alg:permnull}
\begin{algorithmic}[1]
   \State {\bfseries Inputs:} Number of permutations, $B$. Training set labels. Training and s-validation sets of features learned by the DNN.
   \For{$i=1$ {\bfseries to} $B$}
   \State Shuffle the training labels.
   \State Fit a shallow classifier to the shuffled training labels (using the training set of learned-features as inputs).
   \State Generate label predictions using the s-validation set of learned-features.
   \State Evaluate the classification performance of the s-validation predictions against the shuffled training labels.
   \State Save the performance score computed in step 6.
   \EndFor
   \State {\bfseries Output:} Performance metric values saved in step 7.
\end{algorithmic}
\end{algorithm}

In order to understand the rationale behind Algorithm \ref{alg:permnull}, observe that:
\begin{enumerate}
\item The learned-features, used as input in Algorithm \ref{alg:permnull}, represent a transformation from the raw data inputs. Since the DNN used to generate the learned-features is trained with the real labels, and can potentially learn and/or memorize, it follows that the features learned by the DNN can harbor transformed versions of the label structures and/or background structures originally present in the raw input data. Hence, we can still assess if the DNN has learned and/or memorized, by training shallow models with the learned-features.

\item Furthermore, observe that when we train a shallow classifier using the learned-features as inputs and the real labels as outputs, and we evaluate its performance with the s-validation features (as described in step iv above), we have that the observed s-validation scores can potentially capture the contributions of both learning and memorization (since the s-validation features can share both label and background structure with the training features).

\item Finally, observe that when we train shallow models using shuffled labels, as described in steps 3 and 4 of Algorithm \ref{alg:permnull}, we prevent the shallow classifier from learning, since the label shuffling effectively destroys the association between the training features and the labels, making it impossible for the classifier to discover the label structure of the training set. Observe, however, that if the DNN has memorized, then the permutation null distribution will be centered at a classification performance value better than what we would have obtained using random guesses, since the classifier can still recognize background structures in the s-validation features that are similar to background structures that it learned from the training features. On the other hand, if the DNN has not memorized, than the permutation null distribution will be centered around the baseline random guess value, since there is no shared background structure between the training and s-validation features that the shallow classifier can detect.
\end{enumerate}

With the above remarks in mind, we have that by inspecting the value of a given observed s-validation score, and the location of the respective permutation null distribution (relative to the random guess baseline value), we can determine whether the DNN:
\begin{enumerate}[label=\alph*.]

\item Has memorized and learned - in which case the permutation null distribution will be centered at classification performance values (e.g., accuracy) higher than the baseline random guess value (e.g., 0.5, in binary classification problem with balanced proportions of positive and negative labels), while the observe s-validation performance score will likely be even higher than the values achieved by the permutation null.

\item Has memorized, but not learned - in which case the permutation null distribution will be centered at classification performance values higher than the baseline random guess value, while the observe s-validation performance score will likely fall right inside the range of the permutation null distribution.

\item Has learned, but not memorized - in which case the permutation null distribution will be centered around the baseline random guess classification performance value, while the observe s-validation performance score will be better than the random guess.

\item Has neither learned, nor memorized - in which case both the permutation null distribution and the observed s-validation score will be distributed around the baseline random guess value.

\end{enumerate}

Lastly, we point out that while the null distribution generated by Algorithm \ref{alg:permnull} can be used to perform a formal permutation test, comparing the null hypothesis that the DNN has not learned, against the alternative hypothesis that the DNN has learned (irrespective of whether it has also memorized), in practice the corresponding permutation p-values are not really meaningful, since the different s-validation sets could lead to different results, depending on the amount of perturbation used to generate the s-validation sets. As we already mentioned before, the real use of our approach is to compare different DNNs, rather than quantify the absolute amount of learning and memorization achieve by a given DNN.

\section{Illustrations with image data}

\begin{figure*}[t!]
\vskip 0.2in
\begin{center}
\centerline{\includegraphics[width=\linewidth, bb = 0 0 1080 760]{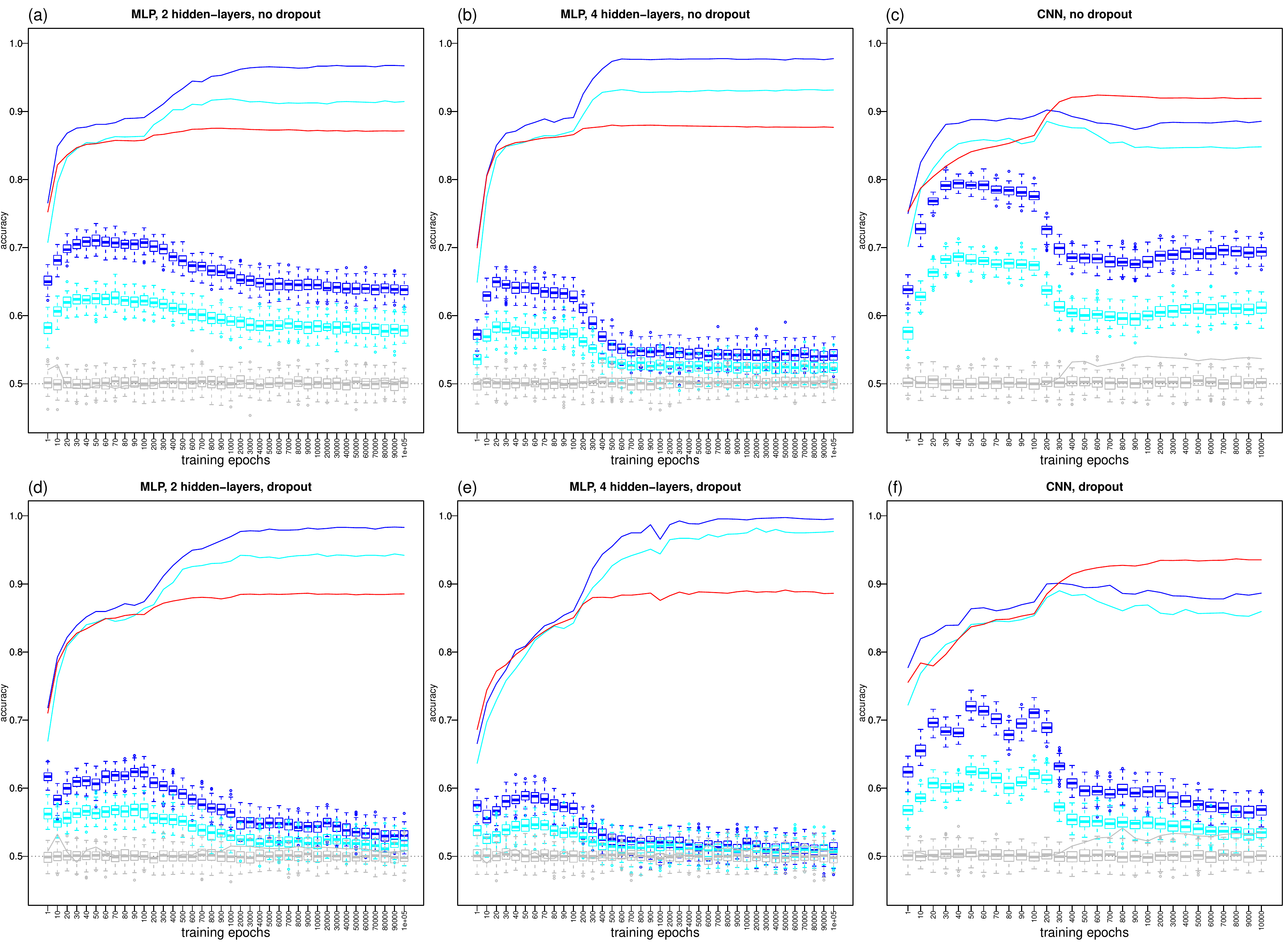}}
\caption{Classification of pullovers vs coats based on MLPs and CNNs. The blue, cyan, and grey boxplots and lines in each panel represent, respectively, the permutation null distributions and observed s-validation accuracies generated by shuffling 5\%, 10\%, and 100\% of the pixels of the original training images. The red curves show the validation set accuracies. The DNNs architecture used to generate the results in panels a-f are described next. Panel a: Input$\rightarrow$Dense(256, relu)$\rightarrow$Dense(16, relu)$\rightarrow$Dense(2, sigmoid). Panels b: Input$\rightarrow$Dense(256, relu)$\rightarrow$Dense(128, relu)$\rightarrow$Dense(64, relu)$\rightarrow$Dense(16, relu)$\rightarrow$Dense(2, sigmoid). Panel c: Input$\rightarrow$Conv2d(32, (3, 3), relu)$\rightarrow$Conv2d(64, (3, 3), relu)$\rightarrow$MaxPooling2d((2, 2))$\rightarrow$Dense(16, relu)$\rightarrow$Dense(2, sigmoid). Panel d: Input$\rightarrow$Dense(256, relu)$\rightarrow$Dropout(0.25)$\rightarrow$Dense(16, relu)$\rightarrow$Dropout(0.5)$\rightarrow$Dense(2, sigmoid). Panel e: Input$\rightarrow$Dense(256, relu)$\rightarrow$Dropout(0.25)$\rightarrow$Dense(128, relu)$\rightarrow$Dropout(0.5)$\rightarrow$Dense(64, relu)$\rightarrow$Dropout(0.5)$\rightarrow$Dense(16, relu)$\rightarrow$Dropout(0.5)$\rightarrow$Dense(2, sigmoid). Panel f: Input$\rightarrow$Conv2d(32, (3, 3), relu)$\rightarrow$Conv2d(64, (3, 3), relu)$\rightarrow$MaxPooling2d((2, 2))$\rightarrow$Dropout(0.25)$\rightarrow$Dense(16, relu)$\rightarrow$Dropout(0.5)$\rightarrow$Dense(2, sigmoid). Here, Dense(U, A) represents a dense layer with number of units given by U, and activation function given by A; Dropout(R) represents a dropout layer with dropout rate R; Conv2d(F, S, A) represent a convolutional layer with number of filters given by F, strides given by S, and activation given by A; and MaxPooling2d(P) represents the max-pooling operation with pool-size P.}
\label{fig:pulloversvscoatsMLPCNN}
\end{center}
\vskip -0.2in
\end{figure*}

We illustrate the application of our permutation approach in binary classification problems employing both MLPs and CNNs, with and without dropout regularization. For each DNN example we extracted 16 learned features and generate null distributions over a grid of increasing training epochs (so that we could investigate the dynamics of how DNNs memorize over extended periods of training). Because, in addition to training the DNN, we also had to generate permutation distributions of shallow classifiers over a long grid of training epochs, we adopted neural networks with relatively small number of hidden layers and small numbers of units per layer (details are provided in the figure captions). Reducing the computational burden was also the reason why we focused on binary classification. Our illustrations were based on the classification of pullovers vs coats using the Fashion-MNIST data-set. We selected the Fashion-MNIST data because their images have higher complexity (compared to MNIST), and we selected the pullovers vs coats comparison because it seemed to provide a more challenging comparison than other possible comparisons. All experiments were run in R\cite{rproject2017} using the \texttt{keras} package\cite{rkeras2017} (a R wrapper of Keras\cite{keras2015}). Except when stated otherwise, all examples were trained in a subset of 2,000 training images (1,000 per class), over 100,000 training epochs for MLPs and 10,000 epochs for CNNs, and optimized using standard stochastic gradient-descent with learning rate 0.01, using mini-batches of 128 images.

Figure \ref{fig:pulloversvscoatsMLPCNN} compares 4 MLPs (containing 2 or 4 hidden-layers, and trained with and without dropout) against 2 CNNs (with and without dropout). The blue, cyan, and grey boxplots in each panel represent the permutation null distributions computed with s-validation sets generated by shuffling 5\%, 10\%, and 100\% of the pixels of the original training images, while the curves with corresponding colors represent the observed s-validation accuracies across the training epochs. The red curve, on the other hand, reports the accuracies computed in an independent validation set. The null distributions were generated by fitting random forest classifiers to 100 random permutations of the trainining/s-validation labels (and using the learned-features as inputs).

\begin{figure}[b!]
\vskip 0.2in
\begin{center}
\centerline{\includegraphics[width=3.2in, bb = 0 0 730 780]{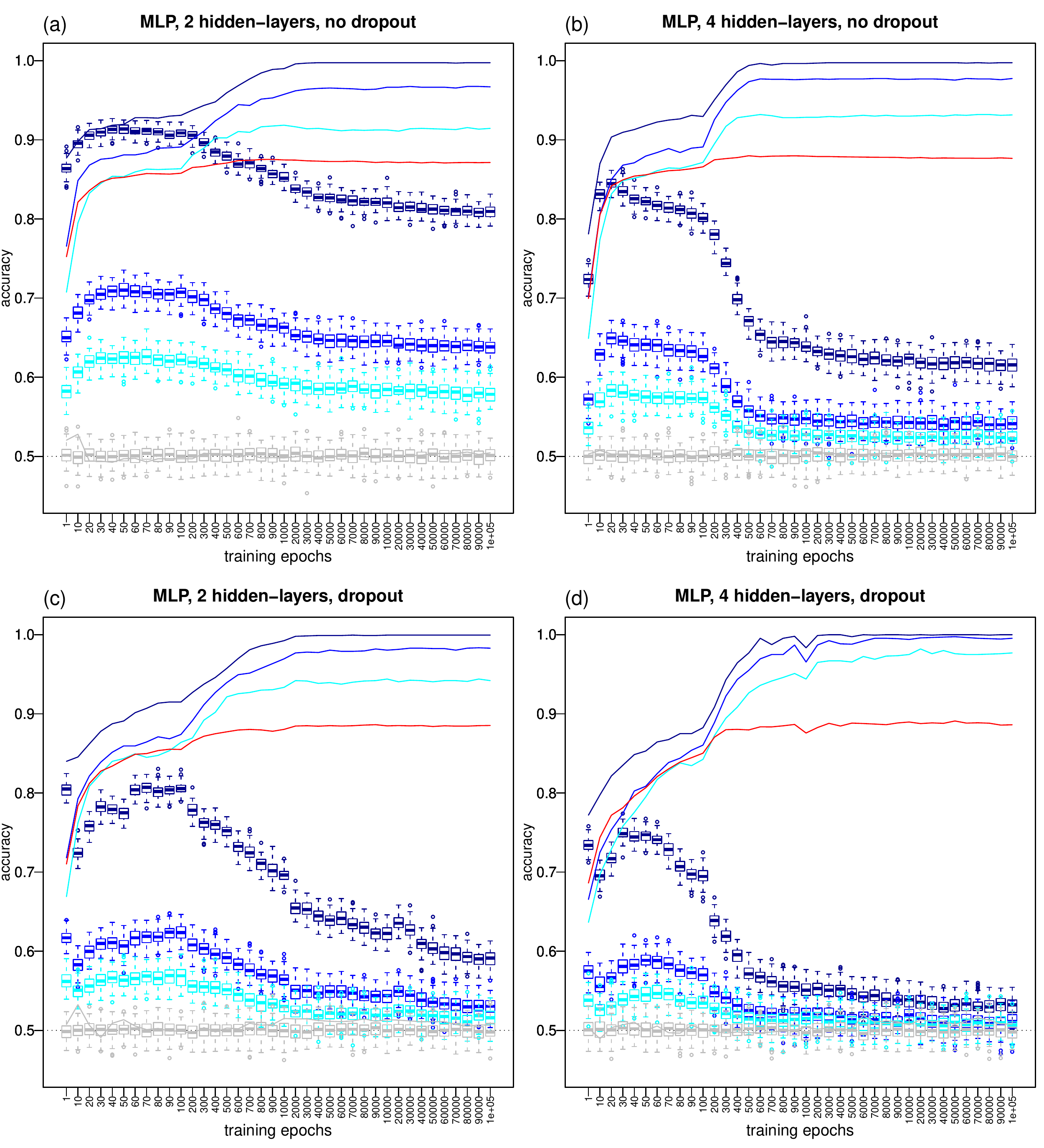}}
\caption{Panels a, b, c, and d show the same examples as in panels a, b, d, and e of Figure \ref{fig:pulloversvscoatsMLPCNN}, but including one extra s-validation set (dark-blue), generated by perturbing only 1\% of the pixels of the training images. Note that while the amount of memorization is considerably higher, we still observe similar patterns.}
\label{fig:pulloversvscoatsMLP}
\end{center}
\vskip -0.2in
\end{figure}

In all examples we observe that the DNNs were learning since the observed s-validation accuracy values (blue, and cyan curves) tend to be well above the range of the corresponding null distributions (blue, and cyan boxplots), just after a few epochs of training. Furthermore, the accuracies in the validation set (red curves) show that the examples are generalizing too. In all examples, we also observe higher amounts of memorization in s-validation sets that share more structure with the training set (note how the blue boxplots dominate the cyan, that dominate the grey boxplots). This is expected since a higher amount of shared structure between the training and s-validation sets makes it easier for the random forest classifier to recognize the background structure it has learned from the training features, and correctly classify the shuffled labels. Observe, as well, that by shuffling 100\% of the pixels of the training images (grey boxplots), we generate a s-validation set that barely shares any structure with the training set. As expected, because the random forest is unable to learn background or label structure in this case, the grey boxplots and the observed accuracies (grey curve) end up being distributed around 0.5, the random guess accuracy value for this problem (note that we have a balanced number of pullover and coat cases). Most importantly, the examples show that increasing the number of hidden layers (in the MLPs) and adopting dropout regularization (in both MLPs and CNNs) helps reduce the amount of memorization. Figure \ref{fig:pulloversvscoatsMLP} shows that these results still hold for MLPs even when adopt a s-validation set generated but shuffling only 1\% of the training image pixels.

The examples in Figures \ref{fig:pulloversvscoatsMLPCNN} clearly illustrate that our permutation approach cannot generally be used to quantify the absolute amount of memorization that is going on in a given DNN, since the location of the permutation distributions depend on the amount of perturbation used the generate the s-validation sets. (The one exception is the case where a DNN is not memorizing at all, since in this case the permutation null distribution will be centered around the baseline random guess value, independent of the amount of perturbation used to generate the s-validation set.) In any case, because all 6 DNNs were trained and evaluated using the same data, we can still evaluate their relative merits. For instance, the MLP in panel e, is clearly memorizing less than the MLP in panel a. Interestingly, the fact that a DNN is memorizing less than another, does not seen to implicate that it will generalize better (note the similar validation accuracies of all 4 MLP examples, in panels a, b, d, and e). Furthermore, even though the CNNs seen to memorize more than the MLPs, they still achieve better generalization. These observations suggest (as pointed out in Arpit et al) that perhaps the capacity of the DNNs is so high that memorization does not seen to hinder their ability to learn and generalize.

\begin{figure}[b!]
\vskip 0.2in
\begin{center}
\centerline{\includegraphics[width=3.2in, bb = 0 0 730 780]{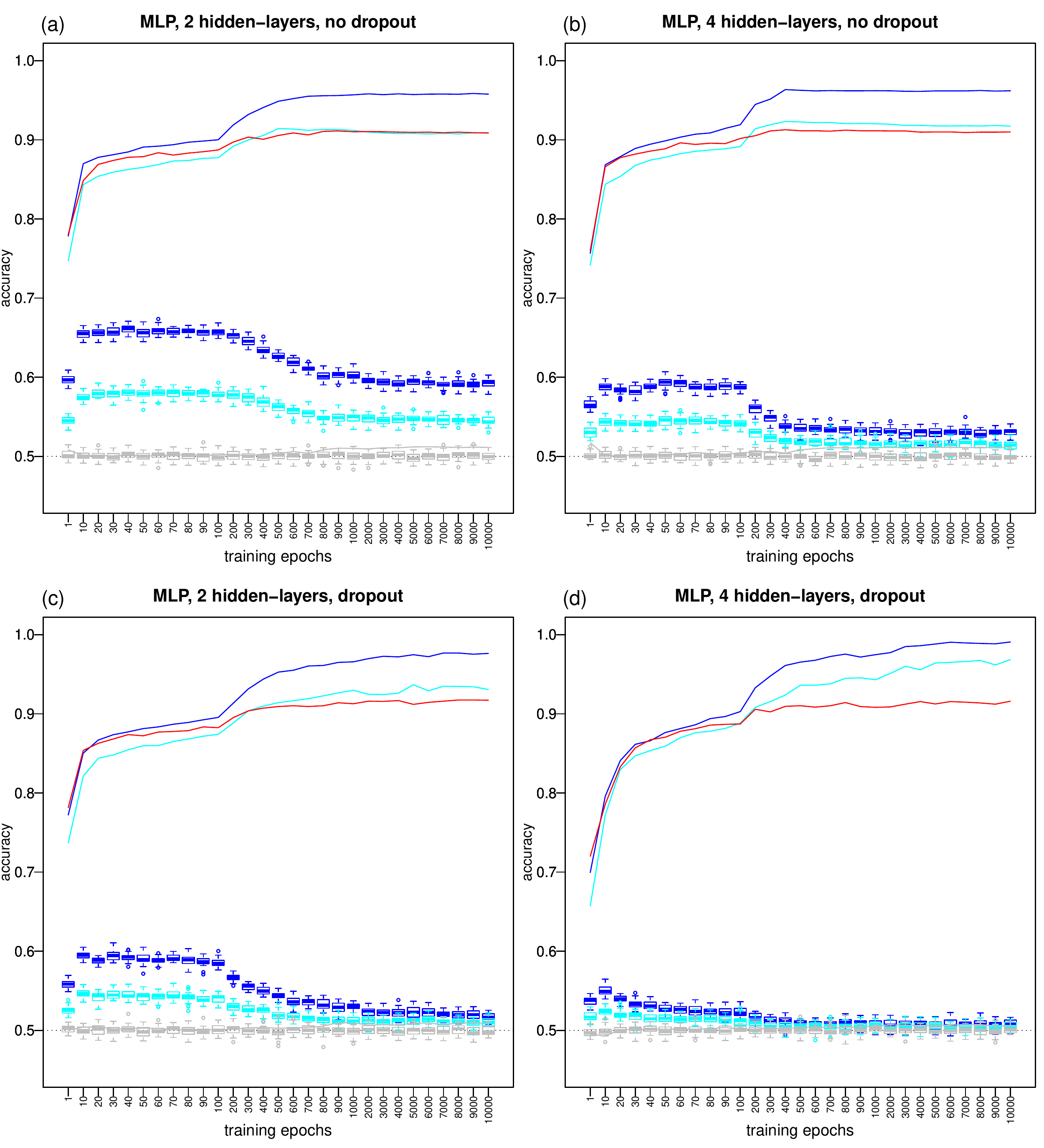}}
\caption{Panels a, b, c, and d show the same examples as in panels a, b, d, and e of Figure \ref{fig:pulloversvscoatsMLPCNN}, but generated using a five fold increase in the number of training images.}
\label{fig:pulloversvscoatsMLPlarge}
\end{center}
\vskip -0.2in
\end{figure}

All the results, so far, were generated using 2,000 training images. Figure \ref{fig:pulloversvscoatsMLPlarge} presents results for the same MLPs trained with 10,000 images. Not surprisingly, we observe a slightly decrease in the amount of memorization and a slight improvement in generalization, but overall we observe the same patterns as before.

Up to this point, all reported examples were optimized using stochastic gradient descent. Quite interestingly, inspection of how the amount of memorization tend to vary as we increase the number of training epochs shows a very interestingly pattern. The blue and cyan boxplots clearly show that the accuracies in the s-validation set tend to quickly increase during the first few hundreds of training epochs, before they tip over and continue to decrease over the many thousand additional training epochs (except for the CNN in Figure \ref{fig:pulloversvscoatsMLPCNN}c). This observation suggests that the DNNs tend to quickly learn a lot of background structure during the first epochs of training, but then as the training goes on, the network is able to keep ``forgetting" about the background structure. Note that this pattern is consistent with recent work on the information-theoretic aspects of learning with DNN\cite{tishby2017}, that has shown that the training of DNNs goes first through a ``memorization" phase, where the DNN quick memorizes the input data during the first epochs of training, followed by an ``information-compression" phase, where the DNN tends to slowly forget about irrelevant structure in the input data by compressing the information via the stochastic relaxation process that naturally takes place in the stochastic gradient descent algorithm. Furthermore, comparison of panels a and b (no dropout case), as well as, panels d and e (dropout case) show that by increasing the number of hidden layers in the MLPs from 2 to 4 we obtain a speed up in how fast the MLPs ``forgets", consistent again with results in\cite{tishby2017}, where it was shown that increasing the number of hidden layers can boost the efficiency of the stochastic relaxation process.

%\footnote{Except for the CNN in Figure \ref{fig:pulloversvscoatsMLPCNN}c, that seems to have started to over-fit the the data, as suggested by the slight decrease in the validation accuracies (red curve) and increase in the null distribution accuracies (blue and cyan boxplots), after 1,000 epochs of training.}

\begin{figure}[b!]
\vskip 0.2in
\begin{center}
\centerline{\includegraphics[width=3.1in, bb = 0 0 730 780]{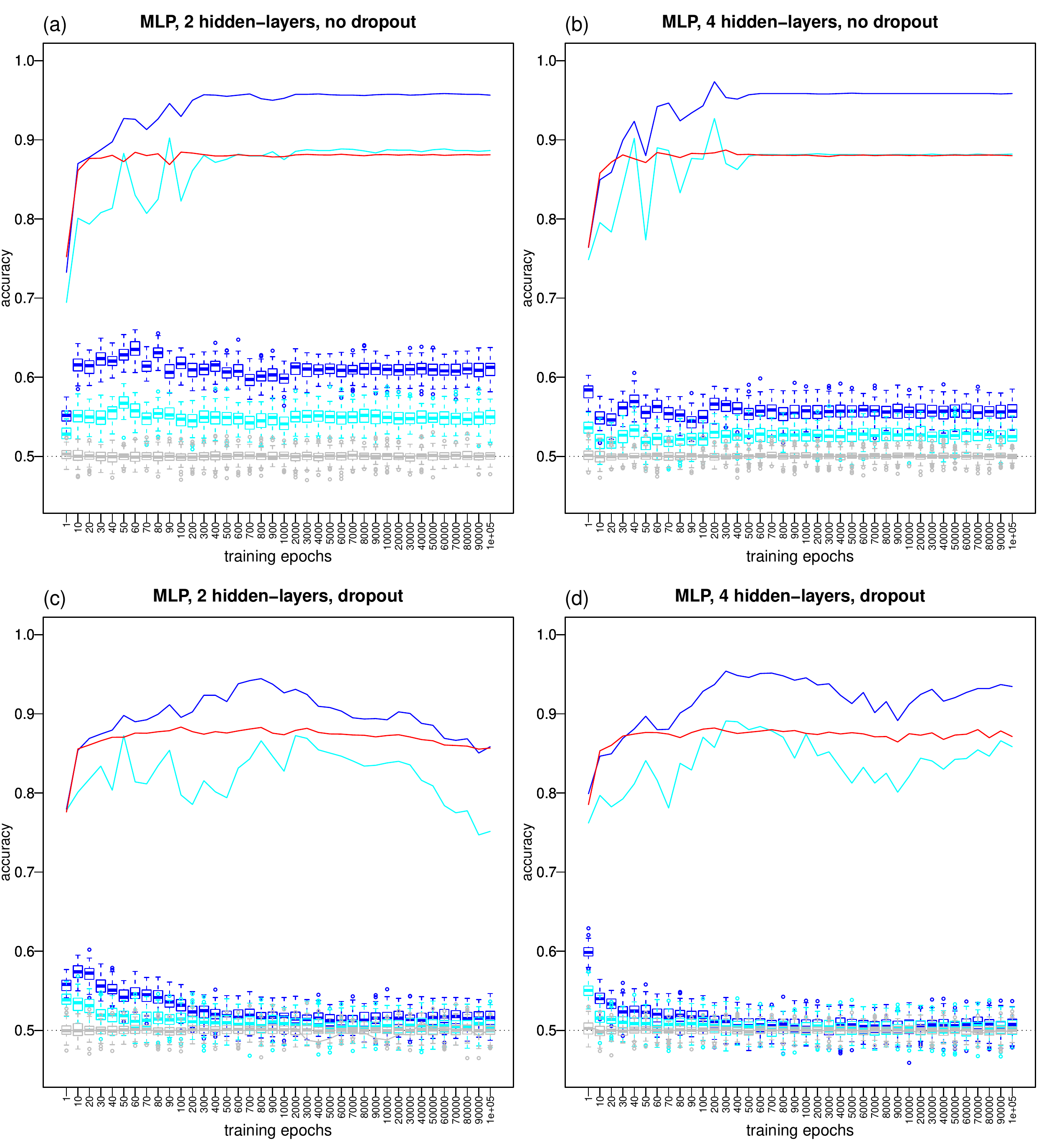}}
\caption{Panels a, b, c, and d show the same examples as in panels a, b, d, and e of Figure \ref{fig:pulloversvscoatsMLPCNN}, but trained with the rmsProp optimizer, instead of standard stochastic gradient descent.}
\label{fig:pulloversvscoatsMLPrmsprop}
\end{center}
\vskip -0.2in
\end{figure}

Figure \ref{fig:pulloversvscoatsMLPrmsprop}, on the other hand, shows that DNNs trained with adaptive learning rate optimizers, such as rmsProp, can follow different memorization dynamics.

\section{Illustrations with Gaussian noise}

Here, we investigate how DNNs learn and/or memorize when trained with pure Gaussian noise as input data. We trained the same DNNs presented in Figure \ref{fig:pulloversvscoatsMLPCNN}, using the same label data, but where we replaced the raw input pixels with Gaussian noise for the training and validation sets. Again, we generated s-validation sets by shuffling 5\%, 10\%, and 100\% of the Gaussian noise values of the training inputs. Hence, even though the raw input data corresponds to pure Gaussian noise, the training and s-validation sets still share a lot of structure (except, of course, when we shuffle 100\% of the values), while training and validation sets are completely unrelated.

\begin{figure*}[t!]
\vskip 0.2in
\begin{center}
\centerline{\includegraphics[width=\linewidth, bb = 0 0 1080 760]{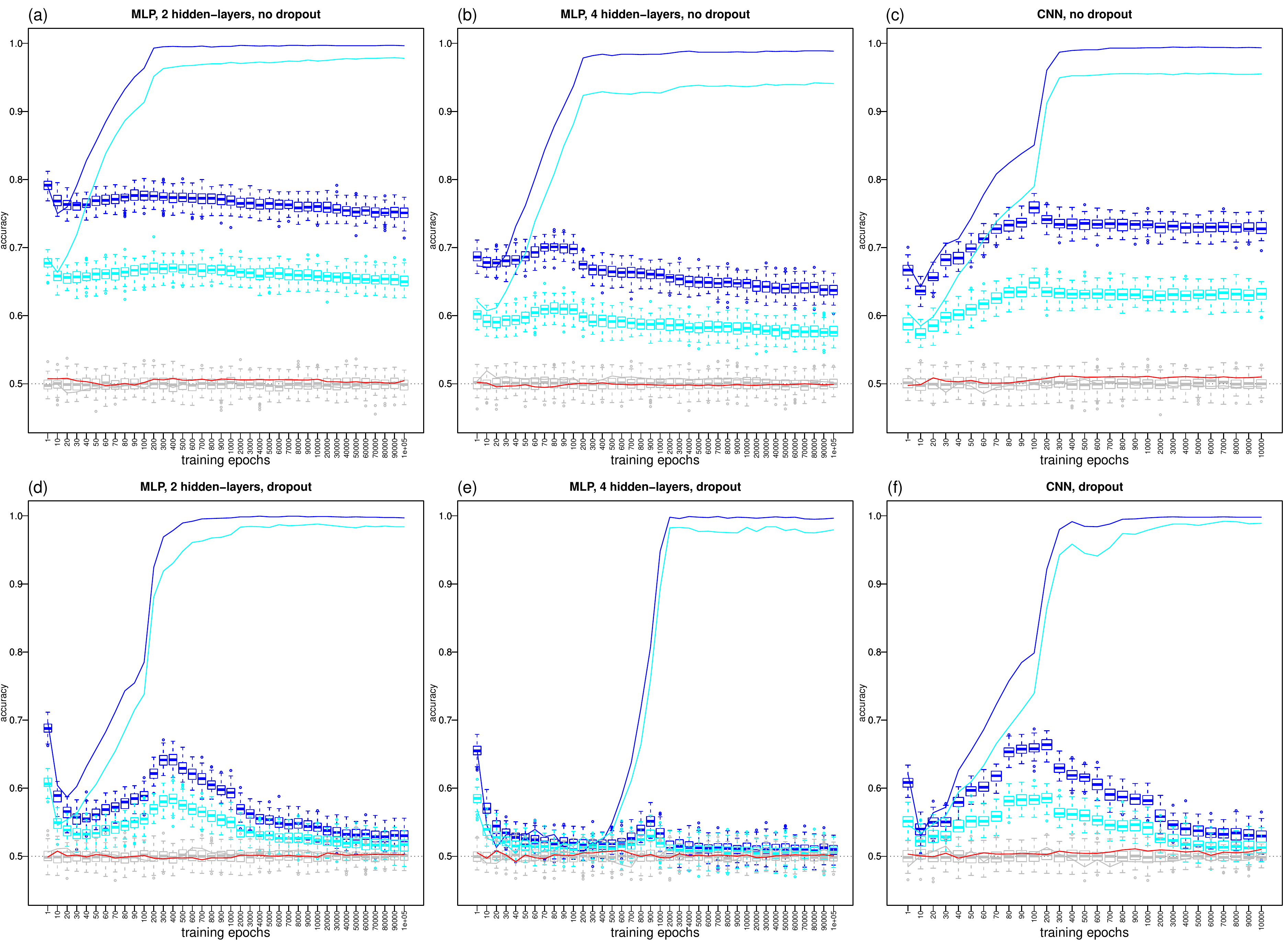}}
\caption{Panels a-f show the same examples as in panels a-f of Figure \ref{fig:pulloversvscoatsMLPCNN}, but where the input data was replaced by pure Gaussian noise.}
\label{fig:gaussiannoiseMLP}
\end{center}
\vskip -0.2in
\end{figure*}

Because it has been shown before that DNNs are also able to reach zero training error when trained with noise inputs (Zhang et al\cite{zhang2017}), we expected that the DNNs would be able to strongly memorize the random gaussian inputs (i.e., we expected that the permutation null distributions would be centered around high accuracy levels). However, we did not expect that the DNNs would be able to learn from the Gaussian noise inputs (i.e., we expected that the observed accuracies of the s-validation sets would fall right inside the range of the respective permutation null distributions). However, as clearly shown by the results reported in Figure \ref{fig:gaussiannoiseMLP}, the DNNs were able not only to memorize, but also to learn with the Gaussian noise inputs. (Note that, in all cases, the observed s-validation accuracies were well above the range of the corresponding permutation null distributions. Note, as well, that he validation accuracies were, as expected, still distributed around 0.5, since none of the random structure that the DNN managed to learn from the Gaussian noise in the training set is shared by the Gaussian noise in the validation set.)

As before, increasing the number of layers and including dropout regularization helped reduce the amount of memorization. Interestingly, the DNNs trained with dropout showed a curious pattern, where during the initial epochs of training they were not learning while showing a decrease in memorization (note how the observed s-validation accuracies tended to be inside the range of the corresponding null distributions), but then they started to quickly memorize and learn (around 40 epochs in panel d, 400 epochs in panel e, and 10 epochs in panel f), before they finally started to ``forget" what they had memorized (around 300 epochs in panel d, 1,000 epochs in panel e, and 200 epochs in panel f), while still retaining what they had learned (note that the observed s-validation accuracies remain close to 1).

These surprising results suggest that the effective capacity of the DNNs is so high, that not only they manage to learn random ``label structure" in pure Gaussian noise, but also they can even learn additional random ``background structure", while learning the ``label structure".

\section{Discussion}

Here, we proposed a novel permutation approach to empirically compare and better understand how distinct DNNs memorize and learn over increasing numbers of training epochs. Our methodology is based on 2 key ideas. The first, is to generate validation sets that share structure with the training set in order to assess memorization. The second, is to assess memorization and learning using shallow models built with the features learned by DNNs trained with real labels, in order the assess the normal contributions of learning and memorization in DNNs trained as usual. Note that while one could still use the s-validation set idea to generate permutation null distributions to detect memorization using the DNNs directly (i.e., by evaluating the predictive performance, on the s-validation sets, of DNNs trained on shuffled labels), this alternative permutation strategy would still only be able assess the memorization behavior that occurs in the absence of learning (which is qualitatively different from the behavior that occurs under normal training conditions, as shown by Arpit et al\cite{arpit2017}).

While our main objective was to introduce the tool, our illustrative examples provide interesting new insights about the few models we investigated here. We believe that researchers interested in understanding how DNNs work will find this tool useful.

%\section*{Acknowledgements}

\end{document}